# Improved Deep Convolutional Neural Network For Online Handwritten Chinese Character Recognition using Domain-Specific Knowledge

Weixin Yang, Lianwen Jin+, Zecheng Xie, Ziyong Feng
College of Electronic and Information Engineering
South China University of Technology
Guangzhou, China
wxy1290@163.com, +lianwen.jin@gmail.com, xiezcheng@foxmail.com, f.ziyong@mail.scut.edu.cn

*Abstract*—Deep convolutional neural networks (DCNNs) have achieved great success in various computer vision and pattern recognition applications, including those for handwritten Chinese character recognition (HCCR). However, most current DCNN-based HCCR approaches treat the handwritten sample simply as an image bitmap, ignoring some vital domain-specific information that may be useful but that cannot be learnt by traditional networks. In this paper, we propose an enhancement of the DCNN approach to online HCCR by incorporating a variety of domain-specific knowledge, including deformation, non-linear normalization, imaginary strokes, path signature, and 8-directional features. Our contribution is twofold. First, these domain-specific technologies are investigated and integrated with a DCNN to form a composite network to achieve improved performance. Second, the resulting DCNNs with diversity in their domain knowledge are combined using a hybrid serial-parallel (HSP) strategy. Consequently, we achieve a promising accuracy of 97.20% and 96.87% on CASIA-OLHWDB1.0 and CASIA-OLHWDB1.1, respectively, outperforming the best results previously reported in the literature.

*Keywords—Handwritten Chinese character recognition; deep convolutional neural network; domain-specific knowledge; hybrid serial-parallel ensemble strategy*

## I. INTRODUCTION

Despite the tremendous works and successful applications of the past four decades [1], handwritten Chinese character recognition (HCCR) remains a major challenge. Numerous methods have been proposed for dealing with confusing characters or cursive handwriting. Among them, classification methods using domain-specific processing technologies such as discriminative feature extraction [2] and discriminative modified quadratic discriminant function (DLQDF) [3] have achieved new levels of performance, attaining test error rates of 4.72% and 5.15%, respectively, on two challenging datasets, CASIA-OLHWDB1.0 and CASIA-OLHWDB1.1 [3],[4]. Under the conventional framework based on the modified quadratic discriminant function (MQDF), preprocessing and feature extraction methods play a crucial role. Such technologies primarily include non-linear normalization methods [5], data augmentation using distorted sample generation [6]-[8], 8 directional feature extraction [9], and similar techniques.

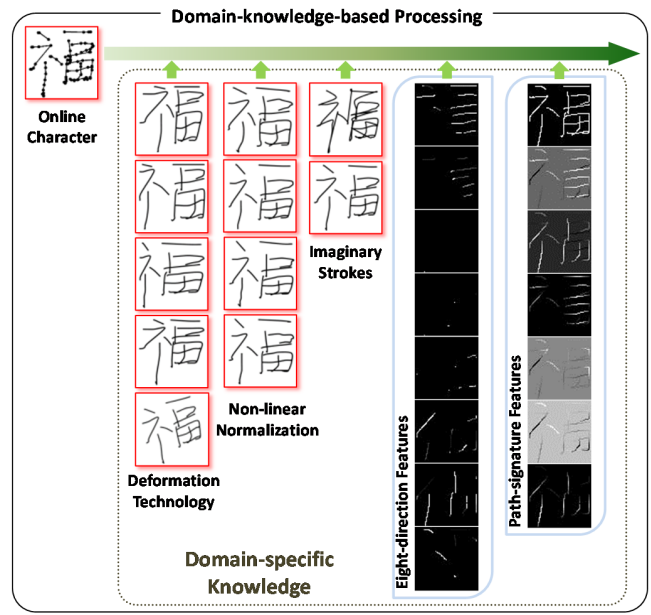

Fig. 1. Adoption of domain-specific knowledge for enhancement of handwritten Chinese character recognition (HCCR) based on deep convolutional neural networks (DCNNs). The first column displays an input character with online strokes. The second column contains five deformation examples, three from Jin [6] and two from Leung [8] and Graham [13]. The third column consists of line density projection interpolation (LDPI) [5], CBA [20], P2DCBA, and P2DLDPF [5]. The fourth column presents the offline image with and without imaginary strokes. The last two columns represent two popular features (i.e., 8-direction features and path-signature features).

All of these traditional domain-specific technologies occupy a significant position in online HCCR.

In recent years, deep convolutional neural networks (DCNNs) have emerged from their noteworthy successes in solving computer vision problems to likewise demonstrate outstanding performance in the field of handwritten character recognition (HCR), beating benchmark performances by wide margins [10]-[12]. The multi-column deep neural network (MCDNN) proposed by Cireşan demonstrates remarkable ability in many applications and reaches lots of near-human performance on handwritten datasets such as MNIST [11]. Graham first proposed a variation of convolutional neural networks (CNNs) called DeepCNet [13], which won first place in the ICDAR 2013 Chinese Handwriting Recognition Competition [14]. By incorporating the path signature feature, DeepCNet produced the best test error rate of 3.58% [13] on CASIA-OLHWDB1.1, which is markedly better than the result of 5.61% from MCDNN [11] or 5.15% from DLQDF [3]. DCNNs have also drawn attention for the applications in the recognition of other Asian character sets such as Hangul [15] and Devanagari [16].

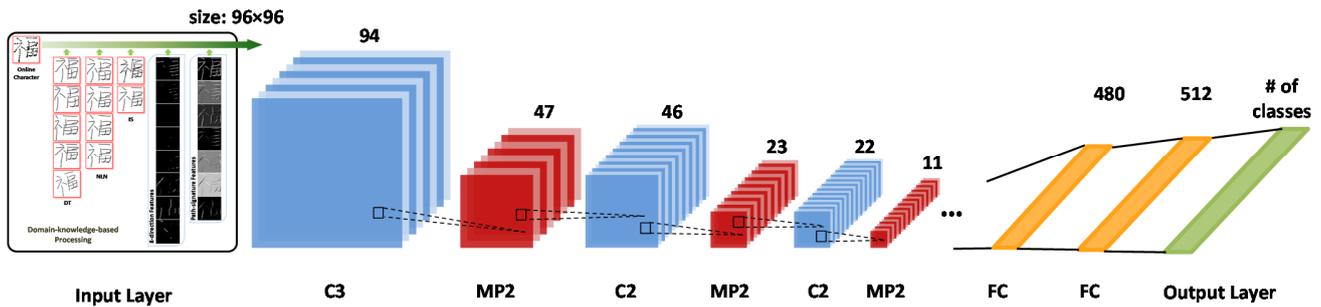

Fig. 2. Illustration of a basic deep convolutional neural network (DCNN) for handwritten Chinese character recognition (HCCR). The settings of the architecture are described in Section II.

However, the performance of current DCNN-based handwritten-character recognizers depends largely on the budget for network scale and time consumption. Accordingly, most research in this field treats the handwritten sample simply as an image bitmap, ignoring some vital domain-specific information that may be useful but cannot be learnt by DCNNs. Therefore, as shown in Fig.1, we propose the idea of incorporating five domain-specific technologies into the baseline DCNN and evaluating the resultant effects both separately and integrally. Furthermore, since each method contributes to a respective DCNN classifier and the results may be complementary with each other, we propose a hybrid serial-parallel ensemble strategy for combining the outputs of all DCNNs. The proposed networks reduce the test error rate to 2.80% on CASIA-OLHWDB1.0 and 3.13% on CASIA-OLHWDB1.1, which are significantly lower than that achieved by other methods as reported in [3],[11],[13], demonstrating that the domain-specific knowledge offers an important means of reducing HCCR error rates.

The rest of this paper is organized as follows: Section II describes our DCNN architecture and configurations. The domain-specific knowledge methods are discussed in Section III, and the experimental results and analysis of the application of these methods are reported on two popular databases in Section IV. Our conclusions are summarized in Section V.

## II. DEEP CONVOLUTIONAL NEURAL NETWORK

Inspired by [11], Graham first proposed a variation of sparse CNNs called DeepCNet [13]. It takes advantages of the sparsity of the input layer and the slow convolutional and max-pooling layers; the slow speed allows for the retention of more spatial information, thus improving the generalization ability. Our DCNNs adopt structures similar to those in [13] but are much thinner and, especially, are enhanced at the first layer with rich domain-specific knowledge. As shown in Fig.2, our DCNN contains six convolutional layers, the first five of which are followed by max-pooling. The size of the convolutional filters is 3×3 in the first layer and 2×2 in the subsequent layers. The convolution stride is set to 1. Max-pooling is carried out over a 2×2 pixel window, with the stride size of 2. Finally, a stack of convolutional layers is followed by two fully connected (FC) layers, which are of sizes 480 and 512, respectively. The number of convolutional filter kernels is much less than that in [13]; ours is set to 80 in the first layer and then increases in steps of 80 after each max-pooling, resulting in a total of 3.8 million parameters, which is much smaller than the 5.9 million used in [13]. For activation functions, rectified linear units (ReLUs) [17] are used for neurons in the convolutional layers and FC layers, and softmax is used for the output layer.

We render the input image into a 48×48 bitmap embedded in a 96×96 grid; thus, the architecture of our network can be represented as follows: M×96×96-80C3-MP2-160C2-MP2-240C2-MP2-320C2-MP2-400C2-480N-512N-Output, where M denotes the number of input channels, varying from 1 to 30 according to the number of different types of domain knowledge incorporated. It is worth noting that the domain-specific knowledge methods presented in Section II play a role only in the input of the network and therefore produce only a minimal increase in the computational burden, while providing considerable flexibility for applicability to other excellent deep models, such as RCNN [18] and Network In Network [19].

## III. DOMAIN-SPECIFIC KNOWLEDGE

### A. Deformation transformation

Deformation technology, in the context of deep neural networks, is used primarily to provide shape variation and to generate a large number of online training data. DeepCNet [13] extends the dataset by applying affine transformations, including global stretch, scaling, rotations, and translations, only using stroke jiggling for generating local distortions. To enrich the data with local diversity, two deformation methods are considered in this paper. One is the one-dimensional deformation transformation [6], which provides adjustable parameters to shrink or stretch parts of the character and create various styles. The other is the distorted sample generation method proposed by Leung [8], which enables both shearing and local resizing.

## B. Non-linear normalization

The NLN method, based on line density equalization, has the goal of shape correction. Line density projection interpolation (LDPI) [5], line density projection fitting [20], and centroid-boundary alignment [20] are three of the most popular NLN methods, and a pseudo-2D normalization strategy [5] is proposed to extend 1D NLN methods. However, according to [15], given that the max-pooling layers in the DCNN can absorb positional shifts, the shape restoration step could be redundant or could even cause loss of information. Moreover, an NLN data preprocessing step is likely to result in loss of diversity, weakened generalization capability, and over-fitting. Thus, a suitable incorporation of NLN into DCNNs is a challenging problem that deserves more attention.

## C. Imaginary stroke technique

Imaginary strokes [21] are pen-moving trajectories in pen-up states that are created to simulate the possible stroke connections formed during rapid cursive writing. Their use has been proved to be effective for HCCR [9],[21]-[23]. Nevertheless, by disturbing the prototype, it could introduce unwanted similarity between characters that are in fact distinguishable. Hence, it is recommended that imaginary strokes should use in conjunction with original strokes to some extent, in which case DCNN could do an efficient learning.

## D. Path signature features

Path signatures, pioneered by Chen [24] in the form of iterated integrals, can be used to solve any linear differential equation and uniquely express a path with a finite length. The path signature feature was first introduced to the recognition of handwritten characters by Graham [13]. Essentially, the zeroth, first, and second iterated integrals correspond to the 2D bitmap, the direction, and the curvature of the pen trajectory, respectively. Depending on the complexity of the HCCR problem, the suggested procedure is to take only these first three iterated integrals into consideration since the following one contributes only insignificance.

## E. 8-directional features

The 8 directional feature [9] is widely used in HCCR for its outstanding ability to express stroke directions. In this technique, features are extracted from each online trajectory point based on eight directions in 2D, and then eight pattern images are generated accordingly. Although the directions can be simplified to 4 or expanded into 16, eight affords a reasonable balance between generalization and precision.

## IV. HYBRID SERIAL-PARALLEL ENSEMBLE OF DCNN

Given that a set of networks offer a diversity of ways to represent different kinds of domain knowledge, using them in combination can be expected to produce better performance (e.g. MCDNN [11]). In view of this, we propose a new DCNN-based ensemble strategy called hybrid serial-parallel (HSP), shown in Fig.3. When a sample enters the HSP-DCNN system, the domain-knowledge-based processing will extract the feature maps for the following DCNNs. The input can pass through a DCNN, which will give the recognition decision if the predicted probability of this DCNN output is greater than a pre-defined threshold $T$ (0.99 in this paper); otherwise it will be sent to the next DCNN with corresponding feature maps from the sample, and so on until either the threshold has been met or it fails to output from the last network. If none of the DCNN output predictions is greater than the threshold, the final recognition decision will be made based on the average output of all the DCNNs. In essence, this proposed HSP classifier ensemble strategy aims to take the advice of the best-qualified expert, while referring the hardest choices to the group to decide as a whole, and we found by experiments (in Section V) that this method gives better results and much less time consumption than just doing a simple voting or averaging procedure.

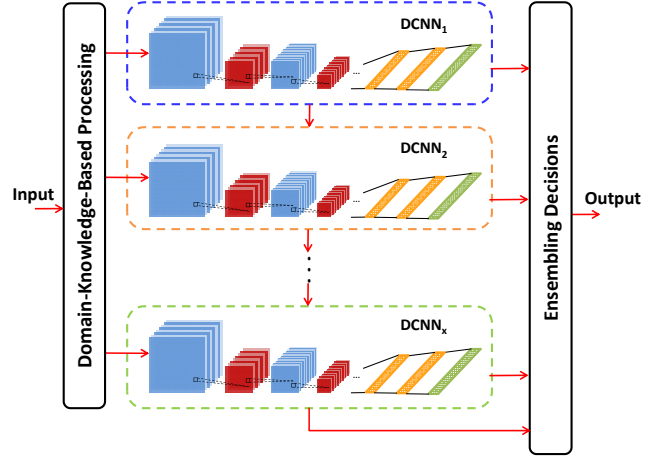

Fig. 3. Illustration of HSP-DCNNs architecture.

## V. EXPERIMENTAL RESULTS AND ANALYSIS

### A. Database and Network Configurations

The databases we used are the CASIA-OLHWDB1.0 (DB1.0) and CASIA-OLHWDB1.1 (DB1.1) [4], which were collected by the Institute of Automation of the Chinese Academy of Sciences. DB1.0 contains 3740 characters of GB1 from 420 writers, 336 allocated for training and 84 for testing, and DB1.1 contains 3755 classes of GB1 from 300 writers, 240 for training and 60 for testing.

For the sake of comparability, we deliberately restricted the single DCNN structure as presented in Section II. For the training stage, a random mix of affine transformations (scaling, rotations, and translations) [13] was taken as the basic elastic distortion (ED) operation. The training mini-batch size was set to 96, and the list of dropouts [25] per weighting layer was experientially set to 0, 0, 0, 0, 0.05, 0.1, 0.3, 0.2. We perform our experiments on a PC with GTX780 GPU and spend about 5 days in training such a DCNN system.

### B. Investigation of Domain-specific knowledge

Because of the outstanding performance reported in [13], we compared the iterated-integrals signature (Sign0, Sign1, and Sign2) in different truncated versions on our network. Note that the Sign0 is directly rendering an online character as an offline bitmap; thus it is regarded as the baseline method. As shown in Fig.4a, the second iterated-integrals signature is cost effective as discussed in Section III. Since most of domain knowledge we adopted is performed based on online information, we thus

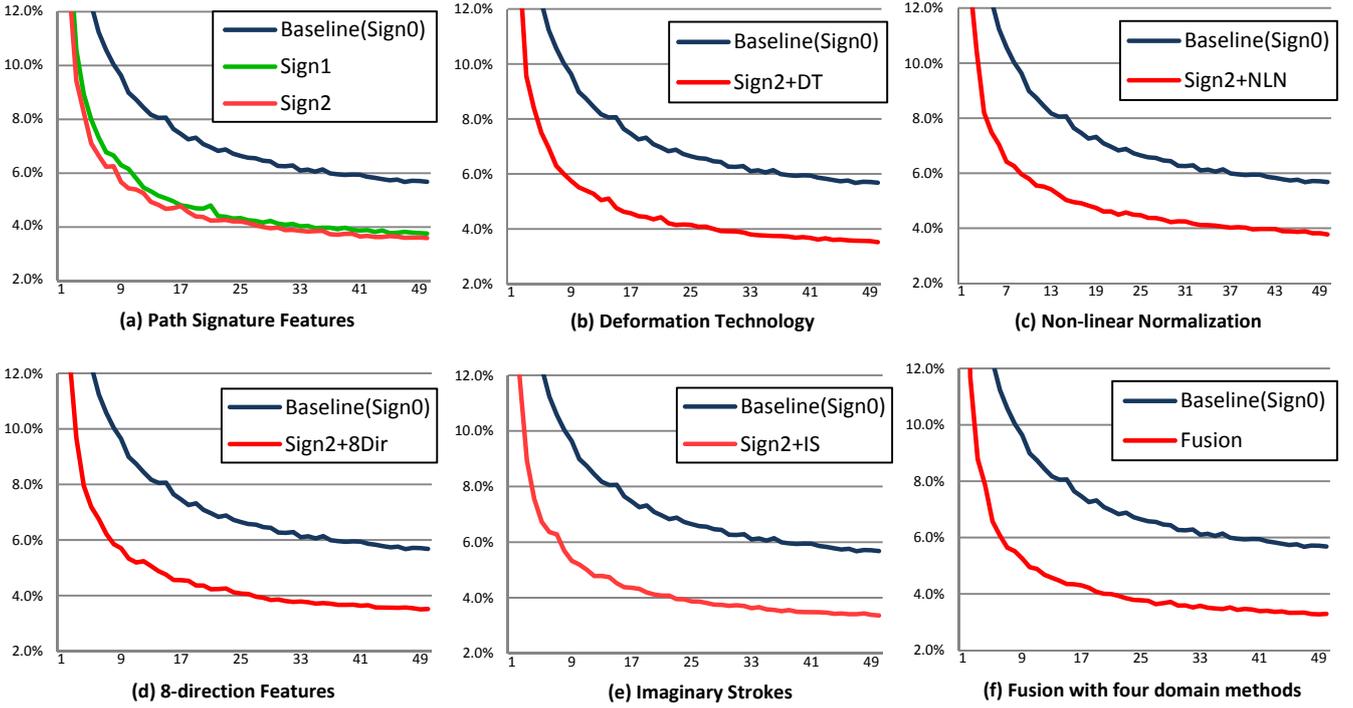

Fig. 4. The performance in error rate of different methods on CASIA-OLHWDB1.0. The x-axis indicates the number (×10⁴) of mini-batch.

TABLE I.    RECOGNITION RATES (%) OF DOMAIN KNOWLEDGE

| Network (3.8 mil.) | Methods | OLHWDB1.0 (38 epoches) | OLHWDB1.1 (62 epoches) |
|---|---|---|---|
| A | Sign0 (**Baseline**) | 94.32 | 93.98 |
| B | Baseline+Sign1 | 96.24 | 95.87 |
| C | Baseline+Sign2 | 96.41 | 96.10 |
| D | Baseline+Sign2+DT | 96.48 | 96.12 |
| E | Baseline+Sign2+NLN | 95.69 | 95.89 |
| F | Baseline+Sign2+8Dir | 96.50 | 96.18 |
| G | Baseline+Sign2+IS | 96.64 | 96.31 |
| H | Baseline+Sign2 +DT+8Dir+IS | **96.72** | **96.35** |

TABLE II.    RECOGNITION RATES (%) OF DCNNs ENSEMBLING

| Database | Single Network | | DCNNs Ensembling (A~H) | | |
|---|---|---|---|---|---|
| | Baseline | Fusion (proposed) | Voting | Averaging | HSP-DCNN (proposed) |
| DB1.0 | 94.32 | 96.72 | 97.16 | 97.19 | **97.20** |
| DB1.1 | 93.98 | 96.35 | 96.84 | 96.86 | **96.87** |

conduct the following experiments incorporated with Sign2 feature maps.

We performed five more experiments (denoted by D through E in Table I) to intensively evaluate the effects produced by the domain-specific technologies embedded in the network C. The experimental results obtained on CASIA-OLHWDB1.0 are shown in Fig.4 and Table I.

A number of interesting points can be observed from the results. First, the deformation technology (denoted DT), which was selected randomly together with the basic ED operations, shows only a slightly better result (Fig.4b), compared with the more obvious improvement in our preliminary experiments on small categories of HCCR. This indicates that although the basic ED has wide coverage of the possible distortions, the additional DT serves to extend it, especially when data are insufficient. Second, the NLN method in [5] (Fig.4c) produces better results than other NLN methods in our preliminary experiments with small categories, but being of inferior performance than network C as shown in Table I, consistent with the limitation noted in Section III, regardless of whether the NLN was applied before or after the ED. Third, with the use of 8 directional feature maps (8Dir) with results as shown in Fig.4d, even though the Sign2 feature maps contain six maps of the first and second iterated-integrals signature, which already include the directional information, the additional 8 directional features are complementary to them, resulting from the statistical effect of directions, where the traditional CNN is helpless. Furthermore, providing imaginary strokes (IS) to a character, resulting in twice as many input maps as those of network C, makes an obvious improvement (Fig.4e) because the network is good at allocating the weights to make a tradeoff. Finally, by combining all of these domain knowledge sources except NLN, we achieve a recognition rate of 96.72% (Fig.4f) on CASIA-OLHWDB1.0, outperforming the 94.32% of the baseline, and indicating a relative error rate reduction of 42%.

*C. Investigation of HSP-DCNNs*

Moreover, we conduct experiments on the ensemble of DCNNs with different strategies in Table II. Note that the result of HSP-DCNNs significantly outperforms that of the single network by a wide margin and is better than other previous widely used ensemble strategies (i.e. voting and averaging), achieving a high recognition rate of 97.20% on DB1.0 and 96.87% on DB1.1, representing relative test error reductions of 51% and 48%, respectively, compared with the baselines. Our final result on DB1.1, indicating a test error rate of 3.13%, outperforms the state-of-the-art results from DLQDF (5.15%) [3] and DeepCNet (3.58%) [13].

TABLE III. COMPARISON OF TIME CONSUMPTION AT TESTING STAGE

| Testing (with GPU) | Single Network | | DCNNs Ensembling (A~H) | | |
|---|---|---|---|---|---|
| | Baseline | Fusion | Voting | Averaging | HSP-DCNN |
| Time per Sample(ms) | 2.40 | 3.50 | 28.10 | 28.08 | 6.22 |

We further evaluate the time consumption of different ensemble methods. As shown in Table III, the proposed HSP-DCNN is much faster than the voting or averaging ensemble strategies. With the HSP strategy, approximately 84% of the test samples can be recognized from the first DCNN with no need for further testing, accounting for its time-saving advantage.

## VI. CONCLUSION

In this paper, we have revealed the potential of domain-specific knowledge for online HCCR in a DCNN-based framework. We discovered that ways can be found to use most of the domain-knowledge-based processing techniques to enhance the DCNN by means of suitable representation and flexible incorporation. The recognition rates of both our best single DCNN and the integrated HSP-DCNN exceed those of the state-of-the-art methods, contributing 20%–22% reductions in recognition error rate compared with previous CNN-based approaches on the CASIA-OLHWDB datasets. In future studies, we plan to investigate additional domain-specific methods to further enhance the DCNN-based framework. In addition, we intend to find a better way of combining all the domain knowledge with proper DCNN architectures.

## ACKNOWLEDGMENT

This research is supported in part by NSFC (Grant No.:61075021, 61472144), National science and technology support plan (Grant No.:2013BAH65F01-2013BAH65F04), GDSTP (Grant No.: 2012A010701001, 2012B091100396, S2013010014240), GDUPS(2011), Research Fund for the Doctoral Program of Higher Education of China (Grant No.: 20120172110023).